\documentclass{bvm} 

\usepackage[nolist]{acronym}

\begin{acronym}
\acro{CLE}{Confocal Laser Endomicroscopy}
\acro{CLE-VS}{CLE video sequence}
\acro{SSL}{self-supervised learning}
\acro{SSIM}{structural similarity index measure}
\end{acronym}

\begin{document}

\selectlanguage{english}

\title{A filtering scheme for confocal laser endomicroscopy (CLE)-video sequences for self-supervised learning}

\titlerunning{A filtering scheme for CLE-VS for SSL}

\author{
	Nils \lname{Porsche} \inst{1},
	Flurin \lname{Müller-Diesing} \inst{2},
    Sweta \lname{Banerjee} \inst{1},
	Miguel \lname{Goncalves} \inst{3},
    Marc \lname{Aubreville} \inst{1}, 
}


\authorrunning{Porsche et al.}

\institute{
\inst{1} Flensburg University of Applied Sciences, Flensburg, Germany\\
\inst{2} University Hospital RWTH Aachen, Department of Otorhinolaryngology, Aachen, Germany
\\
\inst{3} Department of Otorhinolaryngology, Plastic and Aesthetic Operations, University Hospital Würzburg, Würzburg, Germany
\\
}

\email{nils.porsche@hs-flensburg.de}

\maketitle

\begin{abstract}
Confocal laser endomicroscopy (CLE) is a non-invasive, real-time imaging modality that can be used for in-situ, in-vivo imaging and the microstructural analysis of mucous structures. The diagnosis using CLE is, however, complicated by images being hard to interpret for non-experienced physicians. Utilizing machine learning as an augmentative tool would hence be beneficial, but is complicated by the shortage of histopathology-correlated CLE imaging sequences with respect to the plurality of patterns in this domain, leading to overfitting of machine learning models. To overcome this, \ac{SSL} can be employed on larger unlabeled datasets. CLE is a video-based modality with high inter-frame correlation, leading to a non-stratified data distribution for SSL training. 
In this work, we propose a filter functionality on CLE video sequences to reduce the dataset redundancy in SSL training and improve SSL training convergence and training efficiency. We use four state-of-the-art baseline networks and a SSL teacher-student network with a vision transformer small backbone for the evaluation. These networks were evaluated on downstream tasks for a sinonasal tumor dataset and a squamous cell carcinoma of the skin dataset. On both datasets, we found the highest test accuracy on the filtered SSL-pretrained model, with 67.48\% and 73.52\%, both considerably outperforming their non-SSL baselines. Our results show that SSL is an effective method for CLE pretraining. Further, we show that our proposed CLE video filter can be utilized to improve training efficiency in self-supervised scenarios, resulting in a reduction of 67\% in training time.
\end{abstract}

\section{Introduction}
In the process of tumor characterization and outline delineation, tissue samples are taken as targeted biopsies from the most suspicious regions under endoscopic and/or image guidance. However, this procedure is limited by potential negative side effects such as hemorrhage and infection and—importantly for margin confirmation—by sampling error, because only a fraction of the circumference can be interrogated. Margins abutting the skull base (e.g., dura), orbit, brain, the optic nerve, or the internal carotid artery may be inaccessible, where biopsy risks include not only bleeding or infection but also functional loss and/or cerebrospinal fluid leak, which substantially complicates potential reconstruction.
The surgical removal is a complex procedure which aims at complete excision with histologically negative margins (R0 resection) while preserving functionally critical tissue whenever feasible. Functional and essential tissue such as muscles, nerves, and blood vessels may be damaged or removed in the process. A significant reduction in quality of life can occur if limitations arise in speech, sensory perception, swallowing, or breathing functions.

One adjunct to the classical biopsy that has been shown in many recent studies to be used non-invasively and in-vivo in the anatomic region of the head and neck is \ac{CLE}, an optical biopsy technique \cite{sievert-2021-review-cle-hnsscc}. It delivers real-time diagnostic information about tissue structures at a highly magnified scale \cite{sievert-2021-review-cle-hnsscc} and supports intraoperative margin assessment and surgical guidance \cite{VILLARD2022105826}. The challenging aspect of CLE images is in interpreting the data, especially for untrained readers ~\cite{VILLARD2022105826}, where computer-assisted classification can be valuable in the intraoperative field.
At present, CLE is not universally available and is routinely implemented only in a limited number of centers. Moreover, obtaining labeled data is challenging, as reliable correlates are hard to obtain given the margin-sampling constraints described above.
For models with a vast amount of parameters, such as current deep learning models, this data shortage easily leads to overfitting, and hence to a severely reduced performance at inference. 
Few-shot learning methods, recently evaluated for CLE imaging~\cite{aubreville2023-fewshotlearningcle}, offer a potential solution, though classification with limited patient data remains challenging. Another recent strategy in this field is pretraining using \acf{SSL} methods, combined with adaptation to downstream tasks using limited datasets. Given the difficulty of obtaining histopathologic labels in CLE, exploiting unlabeled data becomes particularly valuable. This is especially relevant since ImageNet pretraining introduces a major domain shift, with features poorly capturing the distinct texture and contrast of CLE images.

\begin{figure}[b]
\includegraphics[width=\textwidth]{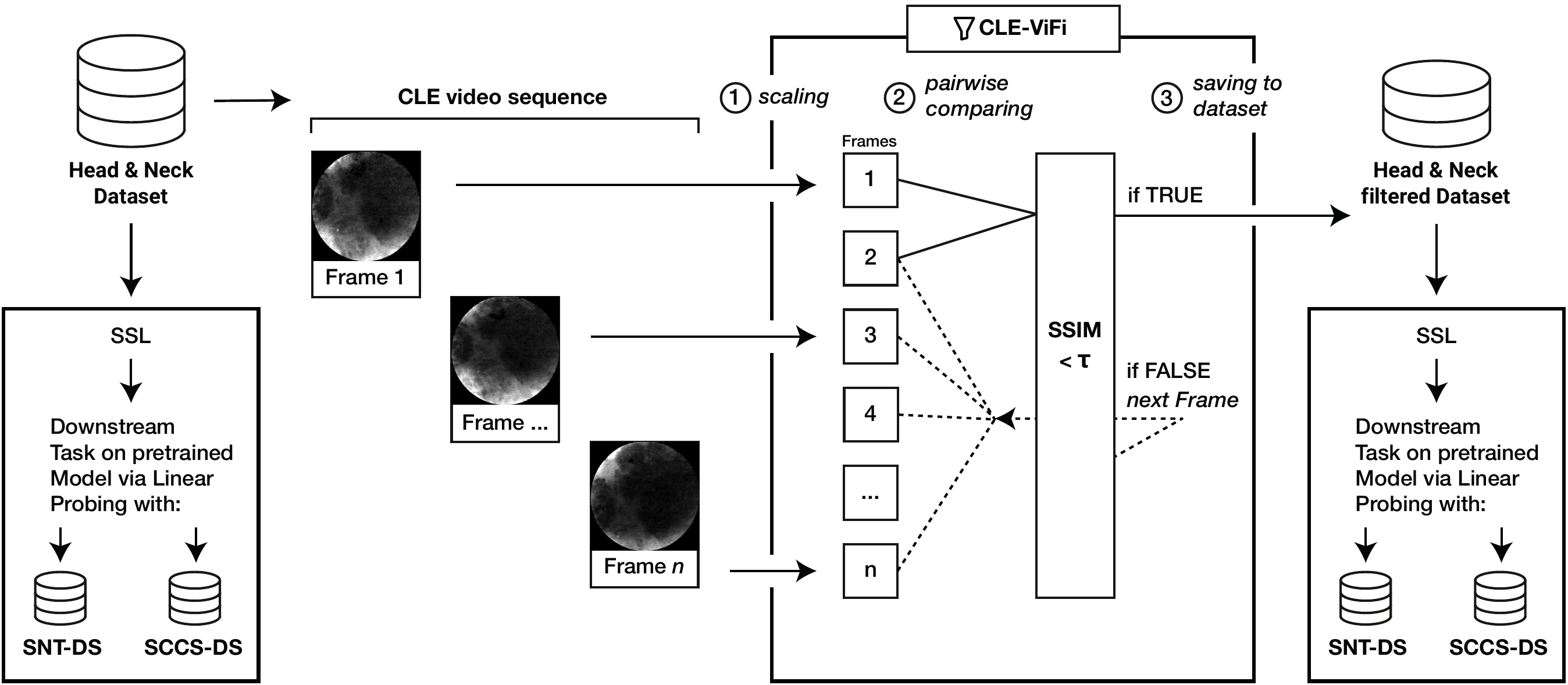}
	\caption{Overview of the approach: We used self-supervised pre-training on a dataset utilizing our proposed CLE video filter (ViFi) and investigated two downstream tasks (SNT-DS, SCCS-DS).}
	\label{fig_CLE_ViFi}
\end{figure}

In this work, we show for the first time that utilizing SSL schemes in pretraining is viable also for the domain of CLE images, even in the context of limited data cohorts. We also propose a novel data filtering scheme specifically tailored for \acp{CLE-VS}, allowing us to reduce the computational time in pretraining while at the same time not sacrificing any performance. In addition, we show in two downstream tasks that our model is able to classify CLE images of sinonasal tumors and squamous cell carcinoma of the skin with significantly increased accuracy than multiple ImageNet-pretrained baseline approaches.

\section{Materials}
We used a total of three datasets from the domain of CLE in this work. The ethics approval was granted by the respective IRBs (243 12 B, 60\_14 B from Universitätsklinikum Erlangen, EK 370/20 from RWTH Aachen, 154/23\_mpz-sc Julius-Maximilians-Universität Würzburg).
For one, we used a dataset of sinonasal tumors (SNT-DS), an anatomical region known for its highly complex and diverse structure. The dataset comprised 42 \acp{CLE-VS} from six patients; only two included both tumor and healthy sequences, while the others contained only healthy tissue, resulting in a total of 6,669 video frames. To enhance the validity of the experiments, we utilized another dataset which consisted of squamous cell carcinoma of the skin (SCCS-DS). It included six patients and 23 \acp{CLE-VS}, totalling 15,740 video frames. In this dataset, one patient had only tumor sequences, while the others had both tumor and healthy tissue sequences.

For self-supervised model pretraining, we used an extended and unlabeled dataset of 458 \acp{CLE-VS}, which we denote the Head and Neck (HAN) dataset. This dataset contains images from various anatomical locations in the field of otorhinolaryngology and oral surgery, including the areas of the vocal folds, squamous cell carcinoma from the oral and sinonasal cavity, as well as from the auricle, nasal cavity, pharynx and larynx. To avoid data bleed in our experiments, both the SCCS and the SNT dataset were not part of the pretraining dataset. The total number of video frames of this HAN dataset was 155,025 (95 GB of raw data), considerably exceeding the scale of the downstream task datasets (SCCS-DS and SNT-DS).

\section{Methods}
\label{0000-sec-latex-vorlage}

\subsection{Self Supervised Learning}
\ac{SSL}, i.e., training of models with a self-created supervisory signal, has recently emerged as a valuable tool in few-shot learning scenarios in medical imaging~\cite{ouyang2022self}. At its core is the idea to utilize large amounts of unlabeled data to pre-train feature extractors, which are subsequently used in downstream tasks using methods such as linear probing or adaptation. This setup helps to regularize the training process, which is key in few-shot scenarios. Caron et al. proposed self‑distillation of vision transformer models by training a teacher and a student on crops of the same image, encouraging the student to mimic the teacher’s predictions across views \cite{caron2021emerging}. This yields robust, view‑invariant representations that generalize well to downstream few‑shot medical tasks, outperforming vanilla pre‑training and providing a simple yet powerful regularization that mitigates overfitting~\cite{caron2021emerging}. \ac{SSL} was shown to significantly benefit from data deduplication \cite{oquab2023dinov2}, which can be attributed to eliminating conflicting signals in the loss function. 

In this work, we train a vision transformer using the original DINO loss \cite{caron2021emerging}. To avoid overfitting, allow for larger batch sizes, and expedite training, we chose the \textit{small} configuration (ViT-small). For the training, we used AdamW as the optimizer and trained the model until convergence, as observed by the validation loss. We then retrospectively selected the model with the best validation loss in training.

\subsection{CLE Video sequence filtering (CLE-ViFi)} 
A core contribution of this work is the inception of a video sequence filtering algorithm. 
A characteristic of CLE video sequences is their high inter-frame correlation, resulting from the relatively steady positioning of the CLE probe during surgical observation. Repeatedly showing nearly identical frames during training is inefficient and, more critically, can introduce conflicting supervisory signals and exacerbate dataset imbalance. As demonstrated by Oquab et al. \cite{oquab2023dinov2}, dataset variability is a key factor for successful \ac{SSL}. In order to reduce the redundancy in the dataset, we propose a tailored video filtering scheme for CLE (CLE-ViFi).

\begin{figure}[b]
\includegraphics[width=\textwidth]{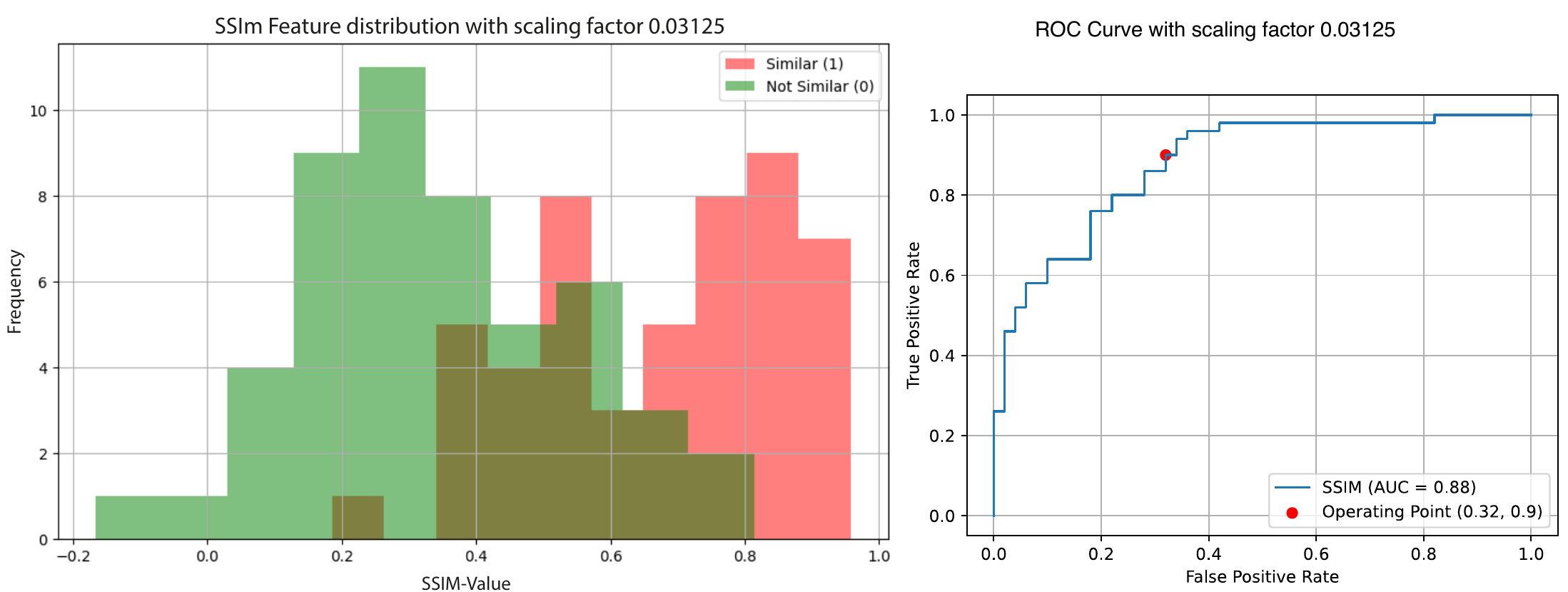}
	\caption{SSIM feature histogram (left) and ROC-AUC curve for SSIM-based thresholding (right).}
	\label{fig_roc_histo}
\end{figure}

Our goal with the CLE-ViFi is to effectively remove all duplicates or near-duplicates in the pretraining dataset. Our work is based on the assumption that the structural similarity of neighboring frames in a sequence can be utilized as a feature for deduplication. However, due to the high noise level in CLE images (see Fig.~\ref{fig_CLE_ViFi}), inter-frame differences in full-resolution data can be strongly influenced by image entropy. Our approach utilizes and combines both these observations.

Using the SNT dataset, we randomly selected 50 reference frames and identified for each a sufficiently dissimilar (i.e., depicting a different anatomical location) and a similar subsequent frame, yielding a total of 100 pairs. We then determine the \ac{SSIM} for all pairs. 

To support the SSIM kernel and reduce sensor-noise within the CLE images, we compared different image scalings for their discriminatory power and found the optimum scaling factor to be 1/32.  We found a sensible operating point for our application at a compromise between false negatives rate (0.1) and false positives (0.32) at a classification threshold $\tau$ of 0.411 (see Fig.~\ref{fig_roc_histo} right). 

Using this frame-based similarity classifier, we now process the entire HAN dataset. We select a first key frame and compare all subsequent frames until we reach a $SSIM < \tau$. This new frame is then added to the filtered dataset and selected as the next key frame. We repeat this until we reach the end of the sequence. This filtering results in the HAN-ViFi dataset, comprising 52,250 non-redundant CLE frames.

\subsection{Downstream Task}
For the downstream task, we employed \ac{SSL}-pretrained models using both versions (HAN, HAN-ViFi) of the dataset. We used linear probing, i.e., the training of a final classification layer on top of the \ac{SSL}-trained feature extractor. We trained the network until convergence, using SGD with a momentum of 0.9 and an initial learning rate of 0.0001 with cosine annealing. 

\subsection{Baselines}
We compared the SSL-based pretraining against four state-of-the-art baseline models. For the first two baselines, we used two convolutional neural networks (CNNs), based on the ResNet-18 and ResNet-50 architectures, respectively, which were pretrained on ImageNet and finetuned on the SNT- and SCCS-Dataset. For the third and fourth baseline, we used the vision transformer base (vit\_base\_patch16\_224) and small (vit\_small\_patch16\_224) which were pretrained on ImageNet-21k and also finetuned on the SNT- and SCCS-Dataset. For all model architectures, we evaluated both full fine-tuning and linear probing. We trained using early stopping based on the validation accuracy using a learning rate of 0.0001 with the Adam optimizer.

\subsection{Cross-Validation}
Given the small size of both of our datasets and the expected data shift between patients, we utilized leave-one-patient-out cross-validation to evaluate all of our model trainings. In this, we always determined one patient to be the hold-out (test) patient in each run. We then split the remainder of the dataset randomly into train and validation on sequence level in a ratio of 80/20, with the condition that at least one sequence of each class was to be present in the training and validation split.

We performed the validation split on sequence level instead of the patient level, because the model training needs learning signals for both, the tumor- and the non-tumor class in each training and validation run. Given the distribution of tumor- and non-tumor cases across patients, a standard split on patient level would not have provided this.
We repeated the cross-validation three times to counter random effects in training (initialization, sampling).

\section{Results and Discussion} 
As shown in Tab. \ref{0000-tab-multi} the pretrained models on ImageNet consistently showed lower performance on the test set of the SNT-dataset and the SCCS-dataset. The SSL models, which were pretrained on CLE data, increased the performance, as measured in the mean accuracy considerably, although not statistically significant (t Test, $p>0.05$). 

We furthermore can see a slight advantage in downstream task performance for the feature extractors trained on the filtered (HAN-ViFi) dataset, leading to the conclusion that the video filtering was not detrimental to \ac{SSL} training. However, by reducing the number of frames in \ac{SSL}-based pretraining by approximately a factor of three, we significantly reduced the training time, as we found model convergence already after 2:27 RTX4090-GPU hours training iterations, compared to 7:23 GPU hours for the unfiltered dataset.

\begin{table}[t]
   \caption{Comparison of the average accuracy $\pm$ standard deviation of the various experiments.}
    \label{0000-tab-multi}
    \begin{tabular*}{\textwidth}{l@{\extracolsep\fill}lllrr}
        \hline
         \multicolumn{3}{c}{} & \multicolumn{2}{c}{Dataset} \\
         Model architecture & Pretraining dataset & Finetuning strategy & SNT-DS & SCCS-DS \\
         \hline
         \multirow{2}{*}{ResNet18} & \multirow{8}{*}{ImageNet (baseline)} & full fine-tuning & $58.87\% \pm $19.11 & $69.70\% \pm $9.72 \\
        & & linear probing & $56.98\% \pm $22.53    & $54.42\% \pm $12.15 \\
         \multirow{2}{*}{ResNet50} & & full fine-tuning & $48.78\% \pm $20.29     & $69.38\% \pm $11.07 \\
         & & linear probing & $66.30\% \pm $29.05     & $61.36\% \pm $16.15 \\
         \multirow{2}{*}{ViT-small} & & full fine-tuning & $59.87\% \pm $25.99     & $68.38\% \pm $11.04 \\
         & & linear probing & $55.93\% \pm $18.39     & $63.79\% \pm $17.42 \\
         \multirow{2}{*}{ViT-base} & & full fine-tuning & $56.48\% \pm $26.45     & $70.06\% \pm $10.93 \\
         & & linear probing & $48.43\% \pm $16.23     & $66.57\% \pm $ 13.92 \\
         \hline
        \multirow{2}{*}{ViT-small} & HAN [ours] & linear probing & $67.20\% \pm $33.83     & $72.53\% \pm $13.19 \\         
         & HAN-ViFi [ours] & linear probing & $\textbf{67.48\%} \pm $34.28    & $\textbf{73.52\%} \pm $12.52 \\
         \hline
    \end{tabular*}
\end{table}

Our investigations support the findings of the current studies in the field of AI research, which predominantly found that SSL-based pretraining can lead to improved results in specialized medical imaging tasks~\cite{huang2023self}. Specifically, our results show that SSL approaches enable more effective use of unlabeled data, which is a major advantage in a field with limited annotated image material. At the same time, our dataset filtering approach demonstrated a reduction in training time without any decrease in test accuracy across both downstream datasets.
Our results show considerable fluctuations, as can be seen in the high standard deviations in Tab.~\ref{0000-tab-multi}. 
This stems from the substantial heterogeneity of the datasets rather than from training instability: the test set, in particular, includes patients with highly diverse and complex anatomical structures, leading to the observed high standard deviations.

\begin{acknowledgement}
Withheld for blind peer review.
The authors would like to acknowledge support by the German Research Foundation (DFG), project number 545049923.
\end{acknowledgement}

\printbibliography

\end{document}